\documentclass[sigconf, screen, natbib=false]{acmart}

\AtBeginDocument{%
  }

\setcopyright{rightsretained}
\copyrightyear{2023}
\acmYear{2023}

\RequirePackage[
  datamodel=acmdatamodel,
  style=acmnumeric,
  ]{biblatex}

\begin{document}

\title{Troubles and Failures in Interactional Language}
\subtitle{Towards a Linguistically Informed Taxonomy}

\author{Martina Wiltschko}
\email{martina.wiltschko@upf.edu}
\orcid{0000-0003-4647-3033}
\affiliation{%
  \institution{ICREA \& Universitat Pompeu Fabra}
  \city{Barcelona}
  \country{Spain}
  \postcode{08018}
}


\begin{abstract}
The goal of this talk is to introduce a systematic research agenda which aims to understand the nature of interaction between humans and artificial conversational agents (CA) (henceforth human-machine interaction, HMI). Specifically, we shall take an explicit linguistic perspective focusing on linguistically defined variables that are known to influence the flow of conversations among humans (henceforth human-human interaction, HHI).
\end{abstract}



\maketitle

\section{Introduction}
The proposed research agenda is built on the premise that there is much to be gained from a comparison between HMI and HHI, both for linguistics as a cognitive science as well as for the design of CAs and the field that underlies it (i.e., conversational artificial intelligence). For the latter, what is at stake is the goal to develop CAs that match the human capacity for conversational interaction. It is the semblance to human conversations that defines one of the core goals of conversational AI: to develop CAs that are capable of natural interaction with humans \cite{wachsmuth_i_2008}\cite{velner_intonation_2020}. This goal, however, raises an important methodological question: What does it mean for HMI to be natural? Currently, the principal metric to evaluate naturalness is how closely HMI resembles HHI. This requires detailed comparison between the human language capacity and the communicative capacity of CAs. To this end, it is essential to develop linguistically informed criteria, which are currently missing despite the fact that the medium of HMI is language. It is one of the goals of this project to fill this gap by using theoretical models of language in interaction. Specifically, I propose to introduce a novel measure of comparison: the use of aspects of language that are dedicated to regulating conversational interaction (henceforth i-language). 

To the best of my knowledge, i-language has never been used to assess the naturalness HMI. Thus, as summarized in Figure 1, I propose to use the comparison of the use of i-language in HHI and HMI to inform conversational AI but also to inform the modelling of human knowledge which underlies conversational interaction. This is summarized in Figure 1. 

The fact that in HMI we cross ontological boundaries in interaction presents us with a novel window into the uniqueness of the human mind. This is critical now more than ever since the boundaries between human and artificial intelligence become more and more blurred with informal reports that ChatGPT might in fact pass the Turing Test.
\begin{figure}[H]
    \centering
    \includegraphics[width=1\linewidth]{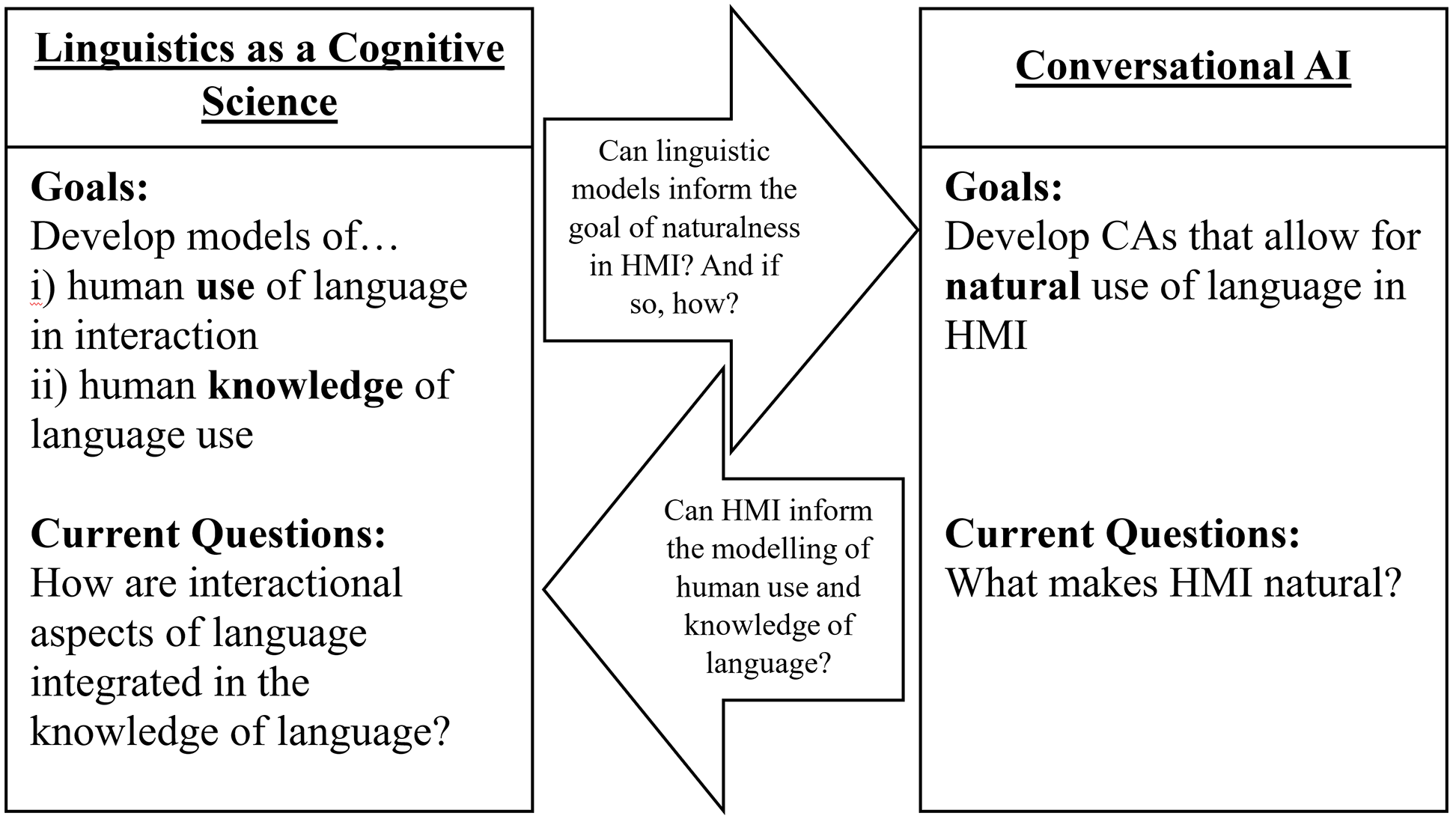}
    \caption{Comparison of \textit{i}-language in HHI and HMI}
    \label{fig:figure1}
\end{figure}
\section{Challenges in HMI: Common Ground, Turn-Taking, and Cooperativity}

What is relevant in the context of the WTF workshop is that current CAs are still limited as interactional partners \cite{van_pinxteren_human-like_2020}. While some of the core ingredients of dialogue systems (automatic speech recognition, natural language  understanding/generation, and text to speech) have improved significantly, conversational skills of CAs are (still) not at the level of human interactants. While their use of representational language expressing propositional content appears to be approaching human capacity, the use of i-language is not. There are several aspects of conversations that are responsible for this status quo, all of which play an important role in HHI and the use of i-language. I discuss each of them in turn. 

\textbf{i) Building common ground} is one of the fundamental driving forces for human conversation \cite{roberts_information_2012} and we rely on our common ground during an ongoing conversation \cite{stalnaker-common-2002}. This is one of the crucial limits of a CA: it cannot participate in common ground building \cite{wang_towards_2021}. The use of i-language reflects assumptions of the speaker about the current common ground, which includes awareness of one’s own knowledge state and assumptions about that of the interlocutor and thus requires a form of Theory of Mind. CAs are mindless interactants and thus common ground building is not something that one would expect from them.

\textbf{ii) Turn-taking }is essential for having successful conversations and in HHI this happens in surprisingly systematic ways. Typical gaps between turns are very short (around 200 ms; \cite{levinson_timing_2015}) and anything that departs from this pattern is interpreted as significant and thus is typically avoided in the normal course of a conversation. While CAs can follow basic turn-taking patterns, they are not performing at the level of a humans: they often interrupt their interactant or conversely take a long time to respond. Thus, for the human interactant, “the conversation feels stilted” \cite{skantze_turn-taking_2021}.

\textbf{iii) The principles of cooperativity} are assumed to guide human
conversationalists in the interpretation of non-literal content (such as metaphors and irony) \cite{cole_logic_1975}. Along with default preferences for agreement and contiguity it has also been shown to be essential in regulating the distribution of i-language: i-language is sometimes used to mark violation of this principle \cite{sacks_preferences_1987}. Though its use requires the ability for inferencing, which is, again, something that CAs may lack (barring recent advances in GPT).

\begin{figure}[H]
    \centering
    \includegraphics[scale=0.15]{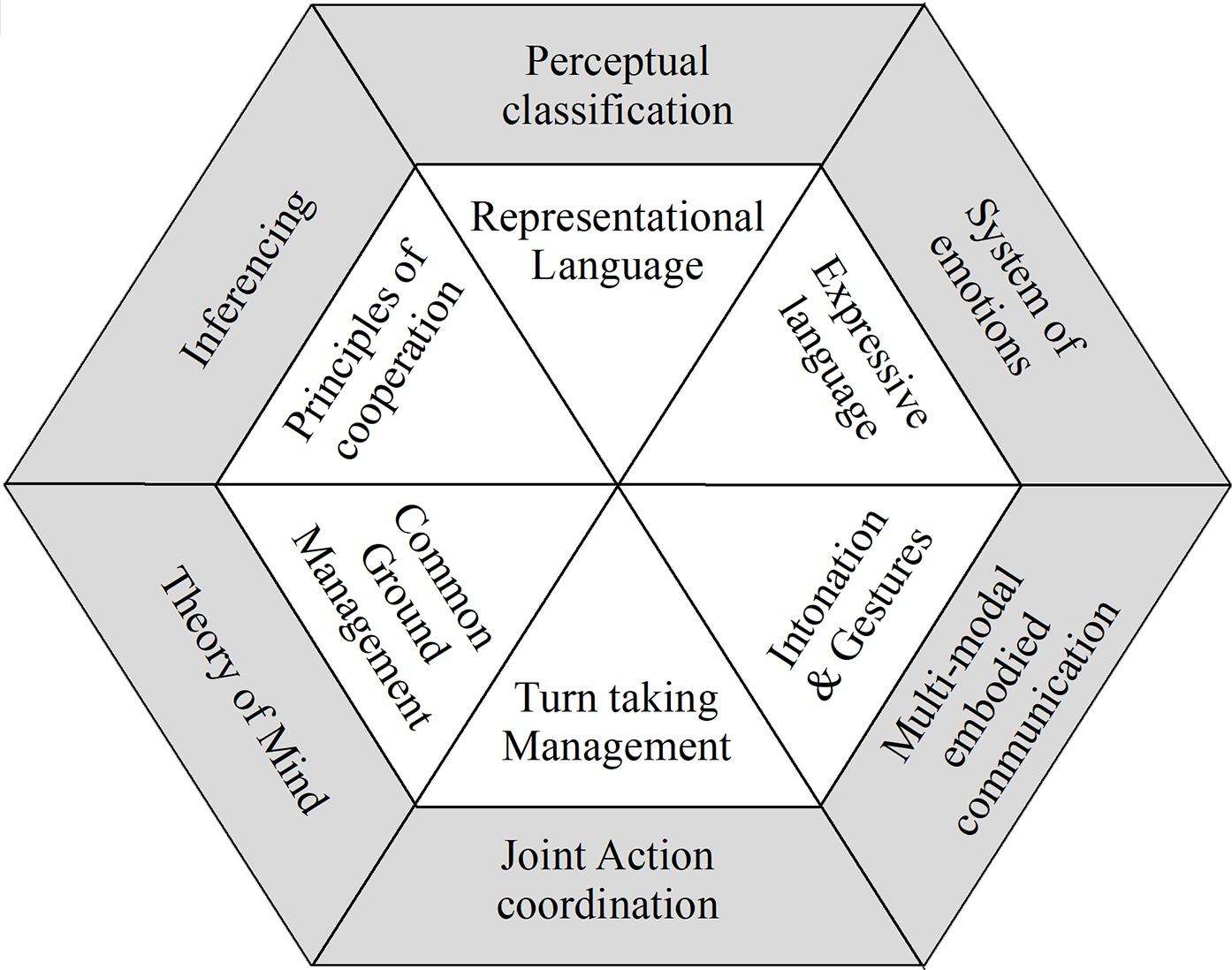}
    \caption{Architecture for human knowledge and use of \textit{i}-language}
    \label{fig:figure2}
\end{figure}
\section{Limitations of Conversational Agents in Mastering i-Language}
There are two more components essential for language in HHI (\textbf{intonational tunes and gestures} and \textbf{emotive language}) which I will not address here, as they do not generalize across all CAs. As illustrated in Figure 2, for each of these properties human interactants have to master the rules of language and conversations (depicted in the white inner hexagon). However, to be properly used, each of these domains has to rely on other general cognitive domains (depicted in the grey outer hexagon). Thus, knowledge and use of language is crucially embedded within (and interacts with) general cognitive capacities (i.e., “hot cognition” ) which AI has thus far not been able to generate to the same extent as representational ‘knowledge’ (i.e., “cold cognition”) \cite{cuzzolin_knowing_2020}. The architecture required for the proper use of i-language contrasts with the typical architecture of a dialogue system which underlie CAs as depicted in Figure 3. (Though, with the advent of ChatGPT and GPT4 it has been claimed that AI might have sparks of artificial general intelligence \cite{bubeck_sparks_2023}, has developed a Theory of mind \cite{kosinski_theory_2023}, and has even been claimed to be conscious by one of Google’s engineers).

\begin{figure}[H]
    \centering
    \includegraphics[scale=0.11]{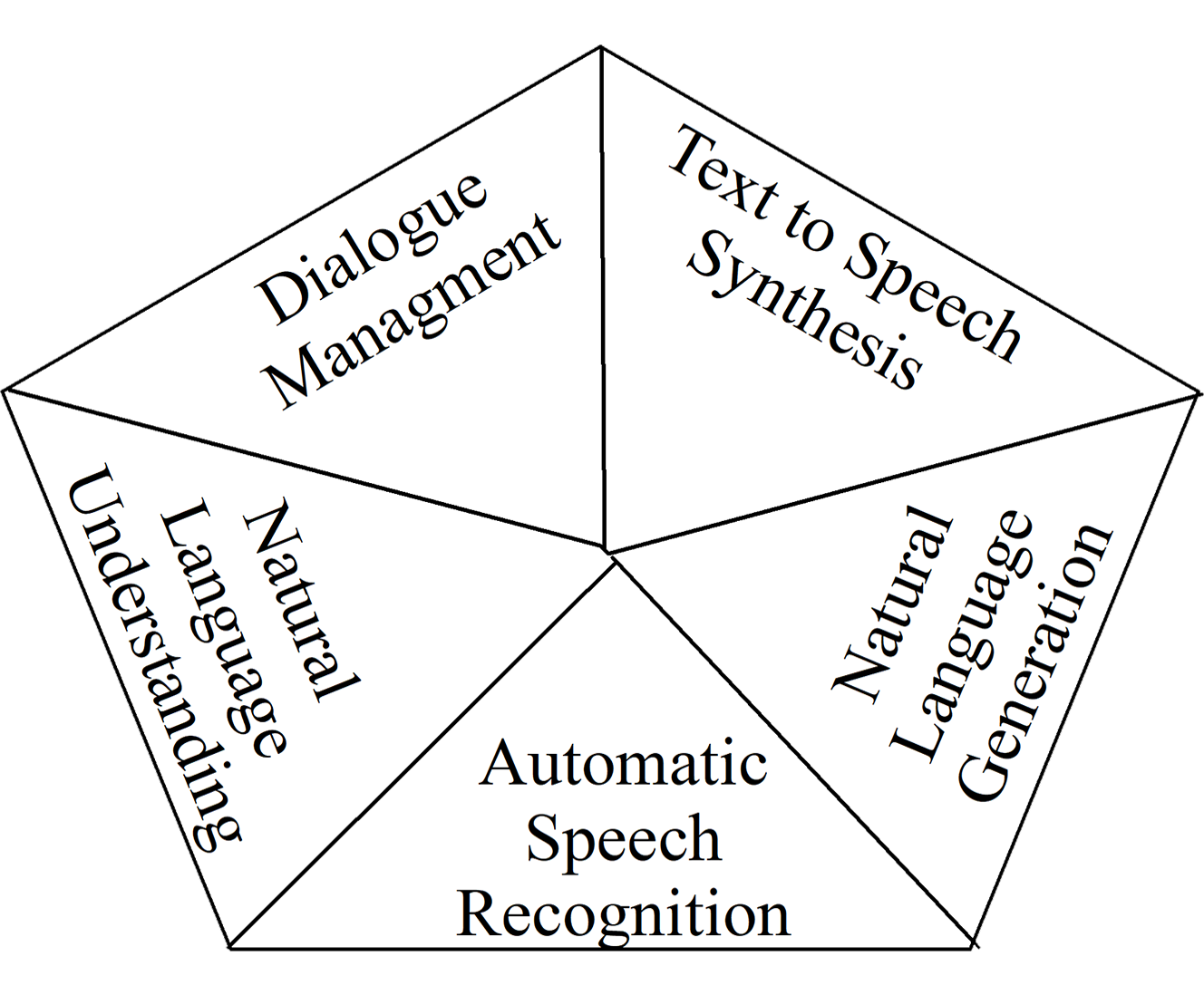}
    \caption{Architecture of a dialogue system}
    \label{fig:figure3}
\end{figure}

The differences in these two figures further highlights differences in the way conversational AI and linguistic theory model language: conversational AI models language via large language models and use of language via dialogue systems. Linguistic theories that are embedded within cognitive science, model not only language and its use, but also the knowledge thereof. Arguably, the use of i-language depends on all of these components and hence CAs are not likely to ever master it. But because i-language is an essential ingredient in HHI, I hypothesize that HMI can never be quite the same. This does however not imply that HMI cannot be natural (just that HHI is not an adequate measure of naturalness). However, the human language faculty is characterized by a unique flexibility to allow adjustments to different interactants, which is part of the creative character that makes human language so unique. It is thus likely that this flexibility will also allow for adjustments of human behaviour in the context of HMI. 

\section{Conclusions and Further Research}
In sum, what I wish to introduce in this talk is a novel taxonomy to explore troubles and failures in HMI that are specifically related to the use of i-language. Specifically, I argue that in order to address current problems in HMI, we need to address the questions in (i-iv) in a way that systematically controls for the components involved in the use of i-language in HHI depicted in Figure 2: \\
\textbf{i) CA production}: Do CAs use i-language adequately?\\
\textbf{ii) CA comprehension}: Do CAs respond adequately to the use of i-language by a user?\\
\textbf{iii) User production}: Do human users use i-language in HMI?\\
\textbf{iv) User evaluation}: How do human users evaluate the use of i-language in HMI?
\printbibliography

\appendix

\end{document}